\pdfoutput=1
\documentclass[11pt]{article}

\usepackage[]{acl}

\usepackage{times}
\usepackage{latexsym}

\usepackage[T1]{fontenc}
\usepackage[utf8]{inputenc}

\usepackage{microtype}

\usepackage{inconsolata}

\usepackage{graphicx}
\usepackage{floatrow}
\usepackage{subcaption}
\usepackage{multirow}
\usepackage{amssymb}
\usepackage[flushleft]{threeparttable}
\usepackage{amsmath}
\usepackage{booktabs}

\usepackage{lipsum}

\usepackage{placeins}
\usepackage{csquotes}
\definecolor{applegreen}{rgb}{0.0, 0.5, 0.0}

\title{BeLLM: Backward Dependency Enhanced Large Language Model\\ for Sentence Embeddings}

\author{
    Xianming Li,
    Jing Li\thanks{\ \ Corresponding author} \\
    \includegraphics[width=0.36cm]{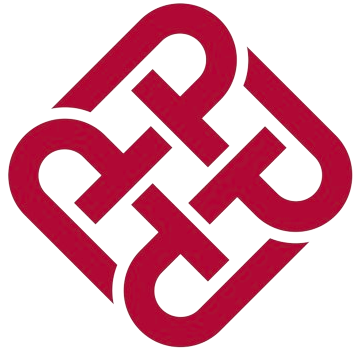} 
    COMP, The Hong Kong Polytechnic University, Hong Kong SAR 
    \\
  \texttt{xianming.li@connect.polyu.hk}, 
  \texttt{jing-amelia.li@polyu.edu.hk}\\
 }

\begin{document}
\maketitle
\begin{abstract}
Sentence embeddings are crucial in measuring semantic similarity. 
Most recent studies employed large language models (LLMs) to learn sentence embeddings.
%Recent studies have proposed using large language models (LLMs) for sentence embeddings. 
Existing LLMs mainly adopted autoregressive architecture without explicit backward dependency modeling. 
%most existing LLMs are built with an autoregressive architecture that primarily captures forward dependencies while neglecting backward dependencies. 
%In this paper, we first present quantitative evidence demonstrating the limited learning of backward dependencies in LLMs. 
%Previous work has highlighted the importance of backward dependencies in improving sentence embeddings. 
Therefore, we examined the effects of backward dependencies in LLMs for semantic similarity measurements.
%To address this issue, 
Concretely, we propose a novel model: \underline{b}ackward dependency \underline{e}nhanced \underline{l}arge \underline{l}anguage \underline{m}odel (BeLLM). 
It learns sentence embeddings via transforming specific attention layers from uni- to bi-directional. 
%Specifically, we found a turning point in LLMs, where surpassing specific LLM layers leads to a significant performance drop in the semantic textual similarity (STS) task, which is crucial for evaluating sentence embeddings. 
%We then extract the layers after the turning point to make them bidirectional, allowing for the learning of backward dependencies.
We extensively experiment across various semantic textual similarity (STS) tasks and downstream applications.
%and draw two findings.
%First, advanced auto-regressive LLMs can benefit from back dependencies for sentence embeddings.
%cannot capture backward dependencies, which are essential in sentence embedding learning.
%Second, 
BeLLM achieves state-of-the-art performance in varying scenarios.
It shows that auto-regressive LLMs benefit from backward dependencies for sentence embeddings.  
\footnote{\ The code is available at \url{https://github.com/4AI/BeLLM}.}
%Extensive experiments demonstrate that BeLLM achieves state-of-the-art performance across various semantic textual similarity (STS) tasks and downstream applications.

\end{abstract}

\section{Introduction}
Sentence embedding is fundamental in natural language processing (NLP). 
% \textcolor{red}{In below, should we focus on the benefits of sentence embedding on semantic similarity measure? Reviewers may think pre-trained models are good enough in representation learning.}
It captures essential semantics in text, benefiting various semantic similarity measurement scenarios \citep{simcse_gao_2021}, such as semantic matching \citep{lu2020deep} and clustering \citep{sbert-nils-2019}.
% affect the performance of various downstream tasks, such as sentiment analysis \citep{zhang2022leveraging}, semantic matching \citep{lu2020deep}, clustering \citep{sbert-nils-2019}, and question answering \citep{yue-etal-2021-contrastive}. 

% The field of NLP research has recently undergone a significant transformation with the emergence of LLMs such as ChatGPT \cite{chatgpt, gpt4} and LLaMA \cite{touvron2023llama2}. 

In previous work, the primary efforts employed smaller-scale bi-directional models \citep{peters-etal-2018-deep, sbert-nils-2019, simcse_gao_2021} to extensively explore the context to learn sentence embeddings.
However, in the paradigm revolution of LLMs and the increasingly large model scales, most advanced NLP models adopted autoregressive (decoder-only) architectures with forward dependency modeling only \citep{touvron2023llama2}.
While some recent efforts used LLMs for sentence embeddings \citep{li2023angle, jiang2023scaling}, limited attention has been paid to studying how backward dependency affects sentence embedding learning in autoregressive architectures.

\begin{figure}[]
    \centering
    \includegraphics[width=1.0\textwidth]{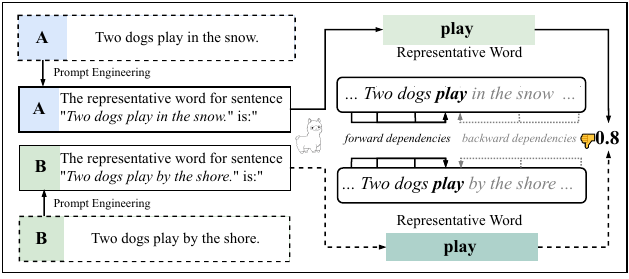}
    %\vspace{0em}
    \caption{
    %A semantic similarity example of LLMs.
    Two sample sentences $A$ and $B$ from STS-B dataset in dashed boxes. 
    %We used previous method \cite{jiang2023scaling} to generate their representitve word ``play'' with prompts in solid boxes. 
    LLaMA predicted $0.8$ similarity for $A$ and $B$ without backward dependency modeling (in grey). The ground-truth similarity is $0.5$ because of differences in the playground in snow and shore.
    }
    %\vspace{-0.5em}
    \label{figure_llm_similarity_case}
\end{figure}

To illustrate the potential help of backward dependencies in sentence embedding, Figure \ref{figure_llm_similarity_case} illustrates two samples from STS-B \citep{cer-etal-2017-semeval} dataset. 
% To enable LLaMA to generate sentence embeddings, we prompt it inspired by \citet{jiang2023scaling}, and LLaMA indicated very high similarity despite the different locations of events.
To enable LLaMA for sentence embeddings, we prompt it inspired by \citet{jiang2023scaling} and observe that LLaMA exhibits a very high similarity despite the different locations of events.
The possible reason is that the uni-directional model LLaMA cannot extensively capture backward dependency, which indicates the relations between the representative word ``play'' and its different playgrounds.
This observation highlights the potential benefits of engaging backward dependencies in LLMs for learning sentence embeddings.

To the best of our knowledge, \emph{our work is the first to extensively investigate the effects of backward dependencies in autoregressive LLMs architectures for sentence embedding learning}.

We start our study with a pilot analysis to quantitatively examine the autoregressive capabilities of LLMs in capturing dependencies. 
It is observed that they are inferior to smaller-scale BERT in these capabilities.
The results suggest the benefits of engaging backward dependencies in LLMs to enhance their dependency-capturing capabilities.

To incorporate backward dependency into LLMs, we propose a novel model, BeLLM, for sentence embedding learning. 
Our core idea is to convert specific attention layers in the transformer decoder from uni- to bi-directional. 
We first conduct a degradation experiment to determine which attention layers should be converted bi-directional. 
It aims to explore the relations between transformer decoder layers 
%(associated with language generation ability) 
and the performance of STS tasks (for semantic similarity measurement). 
%The best STS performance is observed when only converting the last layer to bi-directional and adding more bi-directional layers will render negative effects.  
It is observed that when uni-directional layers exceed a turning point, the STS performance will notablely decrease.
Furthermore, the turning point occurs at the penultimate layer for all LLMs we examined.
%, negatively affecting sentence embedding learning.
% showing the negative effects of these layers for sentence embedding learning.
%Based on this discovery, we extracted the attention layers after the turning point and
We then convert the last layer bi-directional by removing their causal masks.
% It allows all tokens to engage both forward and backward dependencies in attention computation.
% In doing so, BeLLM involves both uni- and bi-directional layers to balance language generation and sentence embedding capabilities.
In doing so, BeLLM involves both uni- and bi-directional layers to balance generation and dependency-capturing capabilities.

% To train BeLLM, we first employ a representative word strategy. It uses BeLLM to generate a representative word for a sentence via prompt engineering.
To train BeLLM, we first employ a representative word strategy to generate a representative word for a sentence via prompt engineering.
% The representative word's embedding hence serves as its sentence's embedding. 
Representative word embedding hence serves as the sentence embedding.
% Its embedding hence serves as the sentence embedding. 
Then, we apply contrastive learning \cite{simcse_gao_2021} to pull embeddings of similar sentences close and push apart those not.

For experiments, we extensively evaluate sentence embeddings learned by BeLLM on various STS tasks.
The main results indicate that BeLLM can significantly outperform previous SOTA on both the standard and the more challenging conditional STS tasks. 
For example, BeLLM achieves $49.74$ Spearman's correlation compared to $47.50$ from prior SOTA \cite{deshpande2023csts}.
It indicates that BeLLM is effective 
%for sentence embeddings 
by introducing backward dependencies into LLMs. 
Then, our 7 downstream tasks experiments suggest that BeLLM's sentence embeddings can benefit various scenarios.
%\textcolor{blue}{
Finally, a case study shows that BeLLM can better measure semantic similarities.
%}

In summary, our contributions are three-fold:

$\bullet$ We explore the dependencies within LLMs and provide quantitative evidence that adding backward dependencies is helpful for sentence embeddings.
% autoregressive LLMs.

$\bullet$ We propose a novel backward dependency-enhanced large language model, BeLLM, to learn sentence embeddings in both uni- and bi-directions. 

$\bullet$ Extensive experiments demonstrate that BeLLM can obtain SOTA results across various STS tasks and downstream applications.

\section{Quantitative Pilot Analysis}
\label{ssec:pilot-analysis}

%We conducted a quantitative analysis to demonstrate the lack of backward dependencies in autoregressive LLMs.
Before introducing BeLLM, we conduct a pilot analysis to explore how advanced LLMs capture dependencies in contexts.
It is driven by the observations that
%\textcolor{blue}{
mainstream LLMs \citep{chatgpt, touvron2023llama2} employ an autoregressive architecture in an uni-directional manner.
%, namely uni-directional network
Intuitively, it lacks the ability to learn backward dependencies \citep{schuster1997bidirectional}, resulting in potential inferiority to capture dependencies compared to its bi-directional alternatives. 
Meanwhile, LLMs possess remarkable emergent abilities \citep{wei2022emergent}, implicitly benefit dependency capturing.  
%leaving uncertainty regarding their capability to capture dependencies. 
We consequently conduct a pilot analysis to quantify whether explicit backward dependency modeling would benefit advanced autoregressive LLMs.
%To address this, we conduct a pilot analysis aimed at quantitatively investigating the dependency-capturing capability of LLMs. 
%}

\begin{figure}[ht]
    \centering
    \includegraphics[width=0.98\textwidth, trim={0 0 0 25}, clip]{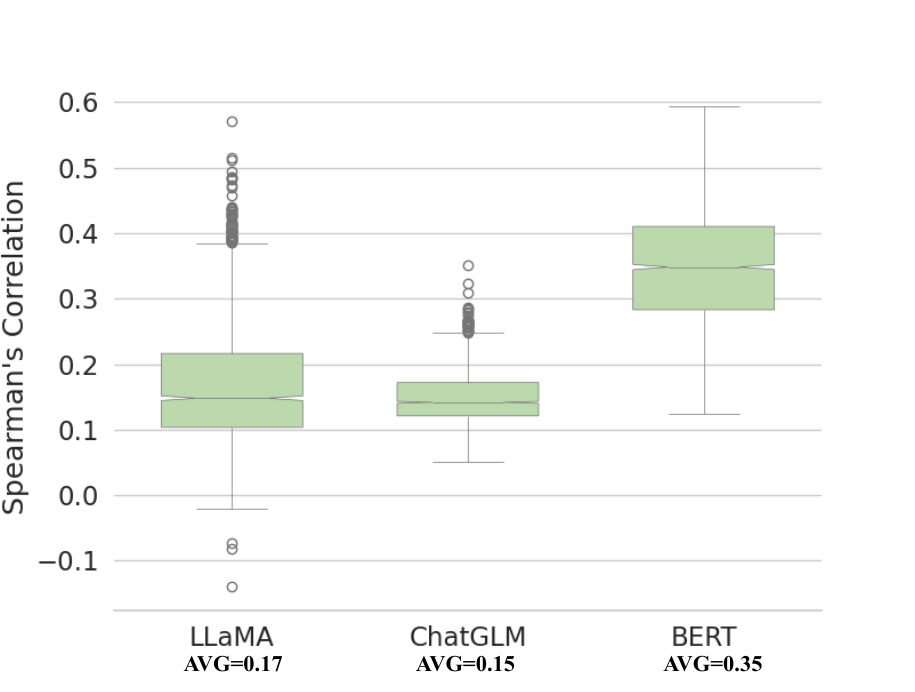}
    \caption{Box plot of the sentence-level Spearman correlation on the STS-B test set.
    % The sentence-level Spearman correlation is calculated by averaging the Spearman correlations between the pivot token and the remaining tokens in the sentence.
    The average sentence-level Spearman correlations for LLaMA, ChatGLM, and BERT are about $0.17$, $0.15$, and $0.35$, respectively.
    }
    \label{figure_quantitative}
\end{figure}

Our analysis is inspired by a phenomenon discovered by \citet{ethayarajh-2019-contextual}.
The study qualitatively shows that bi-directional networks (e.g., ELMo \citep{peters-etal-2018-deep} and BERT \citep{devlin-2019-bert}) demonstrate higher intra-sentence similarity (indicated by word similarities) than uni-directional networks (e.g., GPT-2). %indicating significant word similarities within a sentence. 
%The phenomenon is as follows:
%\textit{
%\textcolor{blue}{bi-directional networks such as ELMo \citep{peters-etal-2018-deep} and BERT \citep{devlin-2019-bert} demonstrate a high intra-sentence similarity, indicating significant word similarities within a sentence. In contrast, the uni-directional network GPT-2 \citep{radford2019language} exhibits lower intra-sentence similarity.}
%}
%This phenomenon was observed, but no explanation was provided. Instead, it was marked as ``unclear'' in \citep{ethayarajh-2019-contextual}.
Here, we further explore this finding from a quantitative perspective.
%We investigate this phenomenon from a dependency perspective. 

Concretely, we experiment on the STS-B test set \citep{cer-etal-2017-semeval} and explore how models in uni- and bi-directional architectures capture dependencies.
For the bi-directional model, we adopt BERT (base) and select the embedding of the ``CLS'' token as the pivot token to represent the sentence embedding. 
For the uni-directional, autoregressive architecture, we employ two representative LLMs: LLaMA \citep{touvron2023llama2} (7B) and ChatGLM \citep{zeng2022glm} (6B). 
They offer two different implementations of autoregressive architectures. 
For them, we choose the last token as the pivot token following \citep{ethayarajh-2019-contextual}.
Based on pivot tokens, we compute their Spearman correlation with the remaining tokens in a sentence to reflect the dependency-capturing capabilities.
%This correlation analysis allows us to assess the dependency-capturing capability of different architectures.
%The sentence-level Spearman correlation from different models is depicted in 
The results are shown in a box plot in Figure \ref{figure_quantitative}. 

%\textcolor{blue}{
The results indicate that BERT shows a higher Spearman correlation, implying its better capability to capture dependencies compared to LLaMA and ChatGLM.
Interestingly, BERT achieves an average score that is about \textit{twice} as high as that of LLaMA and ChatGLM. 
It is possibly attributed to BERT's bi-directional architecture allowing both forward and backward dependency modeling.
In contrast, autoregressive models focus on forward dependencies only.
The results imply the potential benefits of adding backward dependency modeling to autoregressive models for sentence embeddings.

%In contrast, autoregressive models such as LLaMA and ChatGLM focus primarily on capturing forward dependencies, while their capability to learn backward dependencies is comparatively limited.}

% \input{figure_quantitative}

\section{BeLLM}
\begin{figure*}[ht]
    \small
    \centering
    \includegraphics[width=0.9\textwidth]{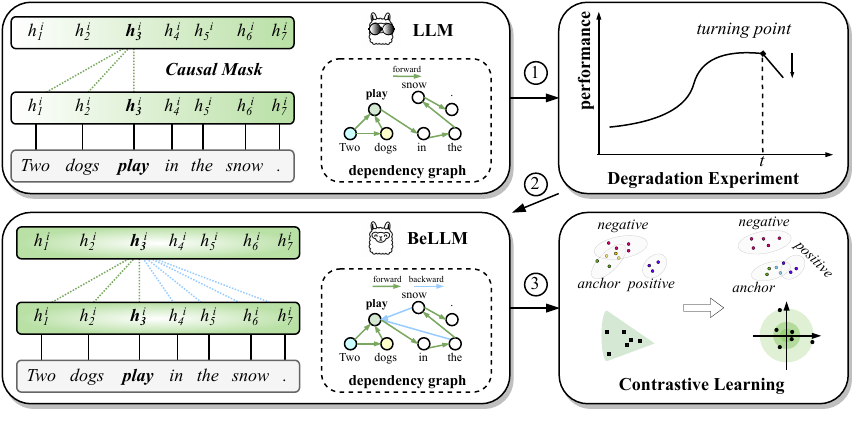}
    \caption{The overall framework of BeLLM. It includes three steps: 1) It first examines how to balance uni- and bi-directional layers with the degradation experiment and finds a turning point. 2) It transforms the attention layers after the turning point from uni- to bi-directional by removing the causal mask. 3) It employs contrastive learning to learn sentence embedding. Here, we visualize the dependencies of the representative word ``play.'' LLM only captures the forward dependencies of ``play'' and BeLLM can capture both forward and backward dependencies.
    }
    \label{figure_framework}
\end{figure*}

The above analysis has revealed the potential benefits of adding backward dependencies to autoregressive LLMs. Consequently, this section will describe our proposed model, BeLLM, with bi-directional attention layers.
%for incorporating backward dependencies into LLMs to enhance the capability of capturing dependencies. 
Figure \ref{figure_framework} shows the overall framework of BeLLM. 
% We will elaborate on the proposed BeLLM in the subsequent sections.
% Subsequent sections are organized following the three steps outlined in the figure. 
Subsequent sections are organized as follows: 
Firstly, we present a degradation experiment to examine how to add backward layers.
%establish a turning point. 
Then, we introduce the architectures of BeLLM, followed by the training methods. 
%and how they help mitigate the anisotropy issue. 
%including the LLM, BiLLM, and Contrastive learning. 
% Finally, we discuss our solution for integrating context modeling results in two directions. 
%Finally, we discuss our solution to the common anisotropy problem in sentence embedding learning. %our solutions to mitigate the anisotropy problem are introduced. 
%Finally, we give a summary.

\subsection{Degradation Experiment}\label{ssec:deg-exp}
\label{model::degradation}
To enable BeLLMs to model backward dependencies, we adopt a straightforward way to turn some uni-directional layers of LLMs into bi-directional ones.
However, uni-directional layers are crucial to LLMs' language generation capabilities
%Since we use prompt engineering to obtain the representative word of the sentence, the generation ability plays a crucial role. 
%Typically, generation ability is associated with the number of layers in LLMs. 
%LLMs with more layers tend to have a stronger generation ability 
\citep{kaplan2020scaling}, which may affect representative word prediction for sentence embeddings (Section \ref{ssec:bellm-training}).

% TODO: ChatGLM-6B
\begin{figure}[ht]
    \centering    \includegraphics[width=0.98\textwidth, trim = {0 0 0 40}, clip]{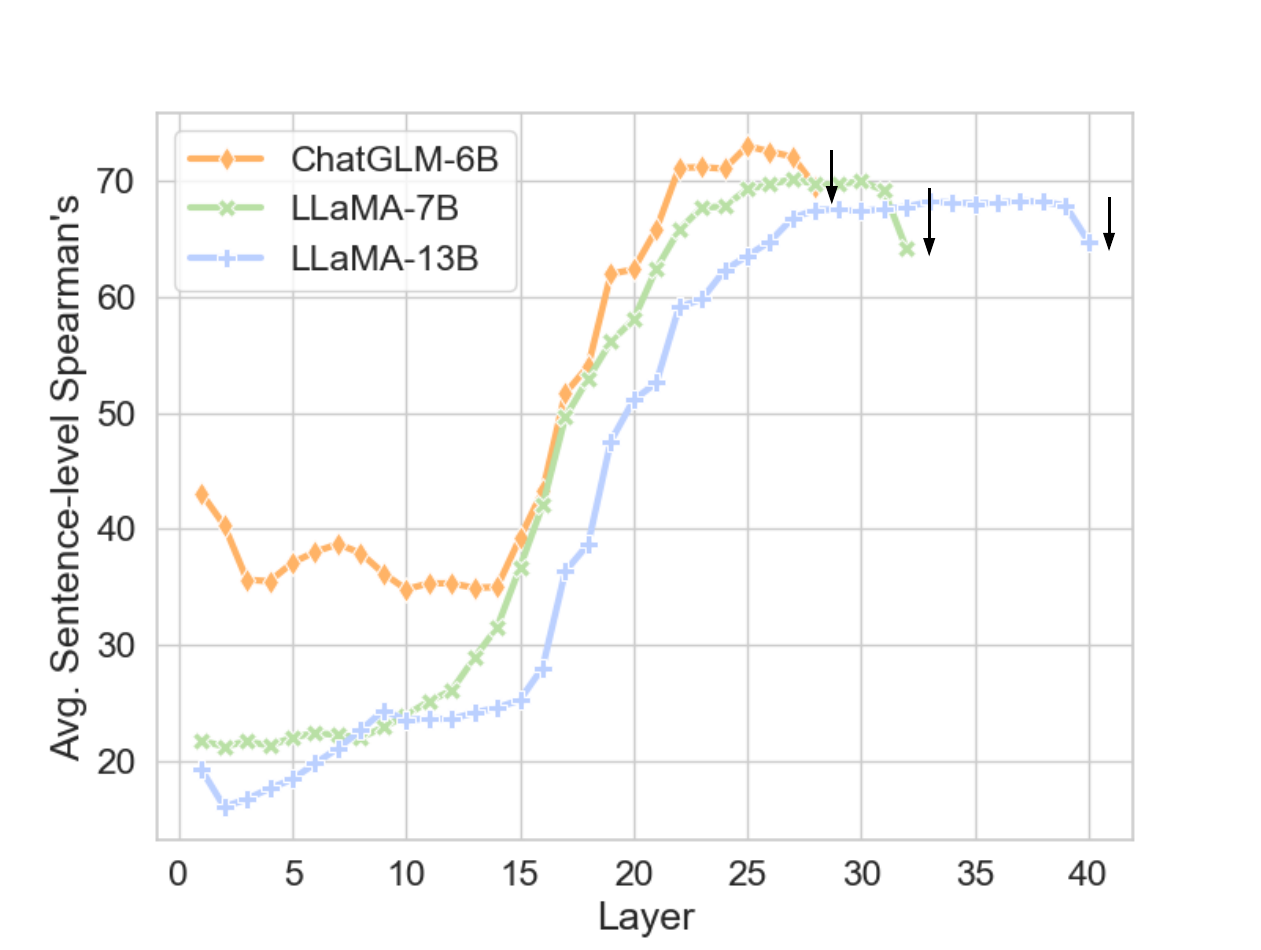}
 \caption{Degradation results on the Standard STS benchmark. X-axis: the number of uni-directional layers. Y-axis: the average Spearman's correlations computed with SentEval \citep{conneau-kiela-2018-senteval}.
% The down arrow indicates the turning point at the last layer.
The down arrow indicates a dramatic performance drop.
}
\label{figure_degradation}
\end{figure}

To practically balance uni- and bi-directional layers, we conduct a degradation experiment to explore the effects of uni-directional layer number on STS performance.  
Here, we gradually removed layers from the last to the first and show the Standard STS benchmark results in Figure \ref{figure_degradation}.
The results present a \textit{S}-shaped curve. It indicates that uni-directional layers are generally helpful to sentence embeddings. 
%overly strong and weak generation abilities fail to produce satisfactory STS performance. 
However, there is a \textit{turning point} at the penultimate layer, exceeding which we observe a consistent performance drop for all three LLMs.
%There is a turning point for generation ability in STS tasks. 
%Once this turning point is exceeded, generation ability no longer positively impacts STS performance; instead, it becomes a negative factor, leading to a dramatic performance drop in the STS tasks. 

Such observations might be attributed to the extreme anisotropy (common-word biases) in the last layer of autoregressive LLMs, also shown in GPT-2's experiments \citep{ethayarajh-2019-contextual}.
%\textcolor{blue}{The figure illustrates a noticeable turning point in all three models: ChatGLM-6B, LLaMA-7B, and LLaMA-13B.} 
%\textcolor{red}{Interestingly, their turning points consistently occur at the penultimate layer. This might be attributed to the extreme anisotropy in the last layer of autoregressive LLMs, similar to the findings on GPT2 \citep{ethayarajh-2019-contextual}.}
Consequently, we convert the last attention layer 
%after the turning point 
from uni- to bi-directional to introduce backward dependencies, yet keeping most layers uni-directional.
%for improving STS.

% \input{figure_degradation}

\subsection{Model Architecture}

BeLLM exhibits layers of auto-regressive LLM and BiLLM (bi-directional LLM) as follows.

%\paragraph{LLM.}
For an input sentence $s$, we first obtain its word embeddings with the following formula:
 
 \begin{equation}
    \small
    \begin{split}
    \mathbf{x} &= \mathrm{Embedding}^{LLM}(t) \\
    t &= \mathrm{Tokenizer}^{LLM}(s).
    \end{split}
\end{equation}

\noindent We then fetch the layers from the first up to the penultimate layer 
%at turning point $t$ from the LLMs, specifically 
denoted as $\mathrm{LLM}^{1:n-1}$ ($n$ indicates the layer number). 
They keep autoregressive architectures to handle language generation effectively and
%and serve as the generation component of BeLLM.
their attention computation is as follows:
\begin{equation}
    \small
    \mathrm{Attn}^{LLM}_i(\mathbf{Q}, \mathbf{K}, \mathbf{V}) = \mathrm{SoftMax}(\frac{\mathbf{Q}\mathbf{K}}{\sqrt{d}} + \mathcal{M}) \mathbf{V} 
\end{equation}
and
\begin{equation}
    \small
    \mathcal{M} = \begin{bmatrix}
     0 & -\infty & -\infty & -\infty & -\infty \\
     0 & 0 & -\infty  & -\infty & -\infty \\ 
     ... & ... & ... & ... & ...  \\ 
     0 & 0  & 0 & 0 &   -\infty \\
     0 & 0  & 0 & 0 & 0 
    \end{bmatrix},
\end{equation}
where $\mathrm{Attn}_i^{LLM}$ is the $i$-th head of multi-head self attention \citep{vaswani2017attention} in LLM. $\mathbf{Q} = \mathbf{W}^q\mathbf{x}+b$, $\mathbf{K} = \mathbf{W}^k\mathbf{x}+b$, $\mathbf{V} = \mathbf{W}^v\mathbf{x}+b$. $\mathcal{M}$ denotes the causal mask. 
It ensures that the $i$-th token is not visible to the $i+1...L$ tokens ($L$ is the input token length) during attention computation, which is crucial to training language generation. 
%This is achieved because the values of $-\infty$ become zero after applying the softmax function.
%These autoregressive layers can retain the generation ability. 
%It is crucial for BeLLM since it uses prompt engineering to generate the representative word of the sentence.
% Then, the embedding of the representative word is used as the sentence embedding. 
% Specifically, we adopt the prompt \textit{The representative word for \{sentence\} is:"} to produce the representative word.

%\paragraph{BiLLM.}
%After turning point $t$, 
Based on the experiments in Section \ref{ssec:deg-exp}, for the last layer (after the turning point), we detach and transform it from uni- to bi-directional to obtain $\mathrm{BiLLM}^{n-1:n}$. 
Its attention is computed as follows:
\begin{equation}
    \small
    \mathrm{Attn}^{BiLLM}_i(\mathbf{Q}, \mathbf{K}, \mathbf{V}) = \mathrm{SoftMax}(\frac{\mathbf{Q}\mathbf{K}}{\sqrt{d}}) \mathbf{V},
\end{equation}
%where $\mathbf{Q} = \mathbf{W}^q\mathbf{x}+b$, $\mathbf{K} = \mathbf{W}^k\mathbf{x}+b$, $\mathbf{V} = \mathbf{W}^v\mathbf{x}+b$.
To engage backward dependencies, we remove the casual mask to turn the last layer bi-directional.
%By removing the causal mask, BeLLM is allowed to learn backward dependencies.

%After obtaining the LLM and BiLLM components, 
At last, we 
couple autoregressive LLM (retaining language generation capabilities) and BiLLM (the bi-directional last layer engaging forward and backward dependency modeling) components. The formula to represent  BeLLM is as follows:
\begin{equation}
    \small
    \mathbf{h} = \mathrm{\overset{\rightarrow}{LLM}}^{1:n}(\mathbf{x}) + \mathrm{\overset{\rightleftarrows}{BiLLM}}^{n-1:n}(\mathbf{x}).
\end{equation}
%where $\mathbf{x}$ is word embeddings. $t$ is the turning point. 

\subsection{Training Methods}\label{ssec:bellm-training}

%\paragraph{Representative Word Strategy.}
%
In the training of BeLLM, we first predict the representative word of a given sentence as the pivot to learn its embedding.
Concretely, we employ a prompt \textit{The representative word for \{sentence\} is:"} for BeLLM to produce the representative word, where \{sentence\} is the placeholder for the actual sentence. Then, the embedding of the representative word serves as the sentence embedding. 

%\paragraph{Contrastive Learning}
Finally, to enable sentence similarity training, we adopt contrastive learning \citep{simcse_gao_2021} to optimize the contrastive objective as follows:
\begin{equation}
    \label{eq_contrastive}
    \small
    \mathcal{L} = - \sum_i \mathrm{log} \frac{
        e^{cos(\mathbf{h}_i, \mathbf{h}_i^+)} / \tau
    }{
        \sum_{j=1}^N \left( 
            e^{cos(\mathbf{h}_i, \mathbf{h}_j^+) / \tau} + e^{cos(\mathbf{h}_i, \mathbf{h}_j^-) / \tau}
        \right) 
    },
\end{equation}
where $N$ is the mini-batch size, $\mathbf{h}_i^+$ and $\mathbf{h}_i^-$ refer to the positive and negative samples of $\mathbf{h}_i$, respectively. $\tau$ is the temperature. $cos(a, b)$ is the cosine similarity function. This way, the embeddings of semantically similar sentences are pulled closer together while dissimilar ones are pushed apart.

%Although adding the backward dependency to BeLLM allows sentence embeddings to exploit the contexts better, 
Moreover, our training methods can potentially mitigate the common \textit{anisotropy problem} in sentence embeddings, 
% This problem
% 
which constrains the embeddings' expressiveness. 
It happens because common words can bias sentence embeddings, rendering the learned embeddings to occupy a narrow cone in the vector space instead of distributing uniformly.
%BeLLM's representative word strategy and contrastive learning can together mitigate this issue.
%Fortunately, our proposed representative word strategy and the use of contrastive learning can mitigate this issue.
The detailed discussions are shown in  Appendix \ref{sec:anisotropy}.

% \textcolor{blue}{
% In summary, incorporating backward dependencies enhances BeLLM's capability to capture dependencies. Additionally, combining the proposed representative word strategy and contrastive learning in BeLLM mitigates the anisotropy problem. These crucial components collectively empower BeLLM to learn high-quality sentence embeddings.
% }

\section{Experimental Setup}
% TODO: report std
\begin{table*}[ht]
\setlength\tabcolsep{1.5pt}
\small
\centering
\begin{threeparttable}
\begin{tabular}{lcccccccc}
\toprule
Model & STS12 & STS13 & STS14 & STS15 & STS16 & STS-B & SICR-R & Avg. \\

\midrule
\midrule
\multicolumn{9}{c}{\textit{Closed-Source Models}} \\
\midrule
% text-embedding-ada-002
openai-ada-002 $\diamondsuit$ & $69.80$ & $83.27$ & $76.09$  & $86.12$  & $85.96$  & $83.17$   & $80.60$ & $80.72$ \\

\midrule
\midrule
\multicolumn{9}{c}{\textit{Unsupervised Models}} \\

\midrule
GloVe (avg.) $\dagger$ & $55.14$ & $70.66$ & $59.73$  & $68.25$  & $63.66$  & $58.02$   & $53.76$ & $61.32$ \\ 
\textbf{BERT} \\
+flow $\ddagger$ & $58.40$ & $67.10$ & $60.85$  & $75.16$  & $71.22$  & $68.66$   & $64.47$ & $66.55$ \\
+whitening $\ddagger$ & $57.83$ & $66.90$ & $60.90$  & $75.08$  & $71.31$  & $68.24$   & $63.73$ & $66.28$ \\
+IS $\ddagger$ & $56.77$ & $69.24$ & $61.21$ & $75.23$ & $70.16$ & $69.21$ & $64.25$ & $66.58$ \\
+CT $\ddagger$ & $61.63$ & $76.80$ & $68.47$ & $77.50$ & $76.48$ & $74.31$ & $69.19$ & $72.05$ \\
+ConSERT & $64.64$ & $78.49$ & $69.07$ & $79.72$ & $75.95$ & $73.97$ & $67.31$ & $72.74$ \\
+DiffCSE & $72.28$ & $84.43$ & $76.47$ & $83.90$ & $80.54$ & $80.59$ & $71.23$ & $78.49$ \\
+SimCSE  & $68.40$ & $82.41$ & $74.38$ & $80.91$ & $78.56$ & $76.85$ & $72.23$ & $76.25$ \\
% LLaMA2-7B $\star$ & $50.66$ & $73.32$ & $62.76$ & $67.00$ & $70.98$ & $63.28$ & $67.40$ & $65.06$ \\

\midrule
\midrule
\multicolumn{9}{c}{\textit{Supervised Models}} \\
\midrule

InferSent $\dagger$  & $52.86$ & $66.75$ & $62.15$ & $72.77$ & $66.87$ & $68.03$ & $65.65$ & $65.01$ \\
USE $\dagger$ & $64.49$ & $67.80$ & $64.61$ & $76.83$ & $73.18$ & $74.92$ & $76.69$ & $71.22$ \\
ConSERT & $74.07$ & $83.93$ & $77.05$ & $83.66$ & $78.76$ & $81.36$ & $76.77$ & $79.37$ \\
SBERT $\dagger$ & $70.97$ & $76.53$ & $73.19$ & $79.09$ & $74.30$ & $77.03$ & $72.91$ & $74.89$ \\
\midrule
\textbf{SimCSE} \\
+RoBERT$_{large}$ & $77.46$ & $87.27$ & $82.36$ & $86.66$ & $83.93$ & $86.70$ & $\mathbf{81.95}$ & $83.76$ \\
+LLaMA $\clubsuit$ & $78.39$ & $89.95$ & $84.80$ & $88.50$ & $86.04$ & $87.86$ & $81.11$ & $85.24$ \\ 
\midrule
BeLLM (ours) & $\mathbf{79.39}_{\ \pm 0.30}$ & $\mathbf{91.04}_{\ \pm 0.15}$ & $\mathbf{86.52}_{\ \pm 0.15}$ & $\mathbf{89.24}_{\ \pm 0.60}$ & $\mathbf{87.43}_{\ \pm 0.3}$ & $\mathbf{89.04}_{\ \pm 0.73}$ & $81.14_{\ \pm 0.80}$ & $\mathbf{86.26}_{\ \pm 0.52}$ \\ 

\bottomrule
\end{tabular}
\end{threeparttable}
\caption{
Spearman's correlation on the standard STS benchmark datasets. 
Higher scores indicate better sentence embedding performance.
%We report Spearman's correlation of the ``all'' setting computed by SentEval \citep{conneau-kiela-2018-senteval}. 
Results marked with $\dagger$ are obtained from \citep{sbert-nils-2019}, those with $\ddagger$ are from \citep{simcse_gao_2021},  those with $\clubsuit$ are from \citep{li2023angle}. 
$\diamondsuit$ indicates the sentence embedding model released by OpenAI, and its results are from \citep{muennighoff2022mteb}. 
Other results are based on reimplementation.
%We refer to the corresponding original papers for the remaining baselines to obtain their results. 
For BeLLM, we report the average scores in five runs with standard deviation (std) expressed as a percentage (\%) after $\pm$.
BeLLM outperforms all baselines significantly on average (p-value < $1\%$, paired t-test). 
}
\label{table_standard_sts}
\end{table*}
%\subsection{Dataset and Evaluation Metrics} 
\paragraph{Datasets.}
We evaluate sentence embeddings on STS tasks following the common practices. It includes the standard and conditional STS as follows.

\vspace{0.5em}
\textit{Standard STS.}  
%\label{exp::standard}
The standard STS (S-STS) benchmark consists of seven STS tasks: STS 2012-2016 \cite{agirre-etal-2012-semeval, agirre-etal-2013-semeval, agirre-etal-2014-semeval, agirre-etal-2015-semeval, agirre-etal-2016-semeval}, SICK-R \cite{marelli-etal-2014-sick}, and STS-B \cite{cer-etal-2017-semeval}. 
%Their test sets are commonly used for evaluating sentence embeddings. 
They contain manual annotations of sentence similarities to test the effectiveness of sentence embeddings.
For training and testing, we follow the prior work \citep{sbert-nils-2019,simcse_gao_2021} to train sentence embeddings on MNLI \citep{williams-etal-2018-broad} and SNLI \citep{bowman-etal-2015-large} datasets. %We then evaluate sentence embeddings 
Then, the trained embeddings are evaluated on the S-STS benchmark datasets.

\vspace{0.5em}
\textit{Conditional STS.} 
To test sentence embeddings with a more challenging setup, we adopt the Conditional Semantic Textual Similarity (C-STS) benchmark \citep{deshpande2023csts}. 
Here, the pairwise sentence similarity is labeled under varying conditions with details in Appendix \ref{sec::csts_intro}.
%Under different conditions, the same pair of sentences can be classified into high and low similarity. 
%The similarity scale ranges from 1 (dissimilar) to 5 (similar)
%Recently, \citet{deshpande2023csts} introduced a challenging 
%Table \ref{table::csts_example} shows an example of the C-STS dataset. 
We follow the setting of C-STS to train sentence embeddings and test them with the C-STS official evaluation API.

\paragraph{Evaluation Metrics.}
For the automatic evaluation of sentence embeddings, we follow previous studies to use Spearman's correlation. 
It compares the ranks of pairwise embeddings and assesses the ranked monotonic relations based on manual annotations.
%for the standard STS evaluation. 
For the S-STS benchmark, we used the SentEval \cite{conneau-kiela-2018-senteval} toolkit 
%to compute Spearman's correlation and 
and reported the results in the ``all'' setting following prior work. 
For C-STS, we reported Spearman's correlation returned by its official evaluation API.

\paragraph{Baselines and Comparisons.} 
For \textbf{S-STS}, we considered widely adopted unsupervised and supervised sentence embedding baselines. 
1) The unsupervised baselines include \underline{GloVe} \citep{pennington-etal-2014-glove} with average pooling,
\underline{BERT-flow} \citep{li-etal-2020-sentence}, \underline{BERT-whitening} \citep{su2021whitening}, and other BERT-based baselines trained with contrastive learning: \underline{IS-BERT} \citep{zhang-etal-2020-unsupervised}, \underline{CT-BERT} \citep{carlsson2020semantic}, \underline{SimCSE} \citep{simcse_gao_2021}, \underline{ConSERT} \citep{consert_yan_2021}, and \underline{DiffCSE} \citep{chuang-etal-2022-diffcse}. 2) The supervised baselines include \underline{InferSent} \citep{conneau-etal-2017-supervised}, \underline{USE} \citep{cer-etal-2018-universal}, \underline{SBERT} \citep{sbert-nils-2019}, \underline{supervised SimCSE}, and \underline{supervised ConSERT}. 
In addition to open-source models, we also compared with closed-source LLM \textit{openai-ada-002} \cite{muennighoff2022mteb}. 

For \textbf{C-STS}, we employed few-shot LLM of \underline{Flan-T5} \citep{chung2022scaling}, \underline{Tk-Instruct} \citep{wang-etal-2022-super}, \underline{GPT-3.5} \citep{chatgpt}, \underline{GPT-4} \citep{gpt4}, and \underline{supervised SimCSE} (fine-tuned) \citep{simcse_gao_2021}.

For baselines' benchmark results, we will report the scores from the original papers and prior work.

\paragraph{Model Settings.}
%In this paper, 
BeLLM employed LLaMA2-7B model \cite{touvron2023llama2} as the backbone.
%Due to the large-scale parameters of LLMs and our limited GPU resources, 
%it is difficult to fine-tune them in full rank. Thus, 
For efficient training, we used the LoRA technique \citep{hu2021lora} for fine-tuning with 
%to fine-tune our BeLLM-based models. 
$lora\_r=32$, $lora\_alpha=32$, and $lora\_dropout=0.1$. We chose the batch size by searching on the values $\{16, 32, 64, 128\}$. The initial learning rate was set to $2e-4$ via grid search on validation data. We set the random seed to $42$ following \citet{simcse_gao_2021} for all experiments.
\section{Experimental Results}

The following will first present the main results with intrinsic evaluation (Section \ref{exp::main}), followed by the extrinsic transfer task results in Section \ref{exp::transfer}.
To provide more insight, we will then analyze the ablation study and case study results in Section \ref{exp::ablation} and \ref{exp::case}, respectively. 
At last, we further examine the intra-sentence dependency of BeLLM to supplement the pilot analysis in Section \ref{ssec:pilot-analysis}.

\subsection{Main Comparison Results}
\label{exp::main}
We first discuss the main results in S-STS and the more challenging C-STS benchmarks.

\paragraph{Standard STS.} 
% We compare our proposed model with widely adopted unsupervised and supervised sentence embedding baselines as follows: 1) the unsupervised models are  GloVe \citep{pennington-etal-2014-glove} with average pooling, BERT-flow \citep{li-etal-2020-sentence}, BERT-whitening \citep{su2021whitening}, and contrastive learning models including IS-BERT \citep{zhang-etal-2020-unsupervised}, CT-BERT \citep{carlsson2020semantic}, SimCSE \citep{simcse_gao_2021}, ConSERT \citep{consert_yan_2021}, and DiffCSE \citep{chuang-etal-2022-diffcse}; 2) the supervised models include InferSent \citep{conneau-etal-2017-supervised}, USE \citep{cer-etal-2018-universal}, SBERT \citep{sbert-nils-2019}, as well as supervised versions of SimCSE and ConSERT. Besides open-source models, we also compare with the closed-source model \textit{openai-ada-002}.

We show  S-STS benchmark results in Table \ref{table_standard_sts} and draw the following observations. 
First, supervised models generally outperform unsupervised models; it suggests that sentence embedding cannot be effectively learned by shallow features, and human supervision can provide positive help.
%the helpfulness of human supervision. 
Second, the LLM-based models perform better than the BERT-based models, implying that larger model scales are helpful in capturing sentence semantics.
%which can be attributed to the larger model scale of the LLMs.
Third, our proposed BeLLM model performs the best in all S-STS datasets.
Specifically, BeLLM achieves a notable $2.5\%$ improvement in average score compared to the previous SOTA SimCSE$_{RoBERTa}$. 
It also consistently outperforms SimCSE$_{LLaMA}$, with a $1.02\%$ improvement in average score. 
These results indicate the effectiveness of incorporating backward dependencies into LLMs for sentence embeddings.
%for improving sentence embeddings.

\paragraph{Conditional STS.} 

\begin{table}[ht]
\small
\centering
\begin{threeparttable}
\begin{tabular}{lc}
\toprule
Model & Spearman's \\
\midrule
\midrule
Flan-T5$_{XL}$ $\dagger$  & $24.80$ \\
Flan-T5$_{XXL}$ $\dagger$  & $29.20$ \\
Flan-UL2 $\dagger$  & $23.20$ \\
Tk-Instruct$_{3B}$ $\dagger$  & $4.90$ \\
Tk-Instruct$_{11B}$ $\dagger$  & $17.10$ \\
GPT-3.5 $\dagger$ & $15.50$ \\
GPT-4  $\dagger$ & $43.60$ \\
\midrule
\textbf{SimCSE} \\
+RoBERTa$_{large}$ (prior SOTA) $\dagger$ & $47.50$ \\
% \midrule
% \multicolumn{2}{c}{\textit{Ours}} \\
% \midrule
+LLaMA & $48.64$  \\
\midrule
BeLLM (ours) & $\mathbf{49.74}_{\ \pm 0.25}\ $ \\

\bottomrule
\end{tabular}
\end{threeparttable}
\caption{Results on the C-STS benchmark. $\dagger$ denotes results from \cite{deshpande2023csts}. 
%LLM baselines are in few-shot settings. 
BeLLM results are the average in five runs. The reported standard deviation (std) value is expressed as a percentage (\%) after the $\pm$ symbol.
BeLLM outperforms all baselines significantly (p-value < $1\%$, paired t-test).
}
\label{table_csts}
\end{table}

To allow sentence embedding evaluation in a more challenging setup, we further employ the C-STS dataset to assess semantic similarities in conditions. 
%to evaluate the performance of BeLLM. 
%To perform the conditional STS task, 
Here, we devise a prompt to enable LLM to summarize the given sentence in one representative word by putting the conditions into the context.
%based on a provided. 
The prompt is as follows: \textit{Given the context \{condition\}, summarize the sentence \{sentence\} in one word:"}, where \{condition\} and \{sentence\} are placeholders for the actual input condition and sentence, respectively. 

The results are presented in Table \ref{table_csts}. 
We observe that LLMs, such as GPT-3.5 and GPT-4, yield inferior results in few-shot settings compared to supervised SimCSE with fine-tuning. 
It implies that C-STS is challenging and cannot be well solved by few-shot LLMs; fine-tuning can helpfully boost the results of a smaller-scale model.
%This observation highlights the importance of fine-tuning LLMs to achieve better performance on conditional STS tasks. 
%As a result, we fine-tune LLaMA and BeLLM specifically for the conditional STS task. 
Among fine-tuned models, the results are consistent with S-STS: larger model sizes and backward dependencies both contribute positively.
Combining their effects, 
%BeLLM performs the best.
BeLLM achieves the SOTA performance.
%Notably, BeLLM outperforms the baselines. 
It demonstrates a $1.10\%$ improvement over SimCSE$_{LLaMA}$ without modeling backward dependency and a remarkable $2.24\%$ gain over the prior SOTA model, SimCSE (RoBERTa$_{large}$). 
%These results provide compelling evidence that the backward dependencies in LLMs are beneficial for improving sentence embeddings and validating the effectiveness of BeLLM.
\begin{table*}[ht]
\setlength\tabcolsep{1.5pt}
\small
\centering
\begin{threeparttable}
\begin{tabular}{lcccccccc}
\toprule
Model & MR & CR & SUBJ & MPQA & SST2 & TREC & MRPC & Avg. \\

\midrule
\midrule

GloVe $\dagger$ & $77.25$ & $78.30$ &  $91.17$ &  $87.85$ & $80.18$ &  $83.00$ & $72.87$ &  $81.52$ \\ 
Skip-thought $\ddagger$ & $76.50$ & $80.10$ & $93.60$ & $87.10$ &  $82.00$ & $92.20$ & $73.00$ & $83.50$ \\
\midrule

Avg. BERT $\dagger$ & $78.66$ & $86.25$ & $94.37$ & $88.66$ &  $84.40$ & $92.80$ & $69.54$ & $84.94$ \\

BERT-CLS $\dagger$ & $78.68$ & $84.85$ & $94.21$ & $88.23$ & $84.13$ & $91.40$ & $71.13$ & $84.66$ \\

IS-BERT $\ddagger$ & $81.09$ & $87.18$ & $94.96$ & $88.75$ & $85.96$  & $88.64$ & $74.24$ &  $85.83$ \\

DiffCSE $\diamondsuit$ & $82.82$ & $88.61$ & $94.32$ & $87.71$ & $88.63$ & $90.40$ & $\mathbf{76.81}$ & $87.04$ \\
\midrule
\textbf{SimCSE} \\
+RoBERTa $\star$ & $83.37$ & $87.76$ & $95.05$ & $87.16$ & $89.02$ & $90.80$ & $75.13$ &  $86.90$ \\
+LLaMA $\clubsuit$ & $90.40$ & $92.90$ & $\mathbf{96.88}$ & $91.57$ & $95.11$ & $95.40$ & $75.13$ & $91.06$ \\
\midrule
BeLLM (ours) & $\mathbf{90.79}_{\ \pm 0.28}$ & $\mathbf{93.43}_{\ \pm 0.50}$ & $96.53_{\ \pm 0.55}$ & $\mathbf{92.01}_{\ \pm 0.32}$ & $\mathbf{95.77}_{\ \pm 0.30}$ & $\mathbf{95.45}_{\ \pm 0.67}$ & $\mathbf{75.48}_{\ \pm 0.59}$ & $\mathbf{91.35}_{\ \pm 0.48}$ \\

\bottomrule

\end{tabular}
\end{threeparttable}
\caption{Accuracy of transfer task results based on different sentence embeddings. 
%of different sentence embedding models (measured as accuracy). 
$\dagger$: results from \citet{sbert-nils-2019}; $\ddagger$: results from \citet{zhang-etal-2020-unsupervised}; $\star$: results from \citet{simcse_gao_2021}. $\diamondsuit$: results from \citet{chuang-etal-2022-diffcse}; $\clubsuit$: results from \citet{jiang2023scaling}. BeLLM's results are the average of five runs. The reported standard deviation (std) value is expressed as a percentage (\%) after the $\pm$ symbol. 
The paired t-test suggests that the average improvements of BeLLM compared to LLaMA are significant, with corresponding p-values of $4.06\%$.
}
\label{table_transfer}
\end{table*}

\subsection{Transfer Task Results}
\label{exp::transfer}
To assess the benefits of sentence embeddings in downstream tasks, we evaluate our model on seven transfer tasks: MR \citep{pang-lee-2005-seeing}, CR \citep{hu2004mining}, SUBJ \citep{pang-lee-2004-sentimental}, MPQA \citep{wiebe2005annotating}, SST2 \citep{socher-etal-2013-recursive}, TREC \citep{voorhees2000building}, and MRPC \citep{dolan-etal-2004-unsupervised}. 
The transfer task results are measured by SentEval \citep{conneau-kiela-2018-senteval} toolkit. 
Following common practices, we train a logistic regression classifier using sentence embeddings as features and used default configurations of SentEval for a fair comparison.

The results are presented in Table \ref{table_transfer}. 
As can be seen, BeLLM achieves superior performance compared to the baselines, obtaining the best results on average and in 6 out of 7 tasks.
%Furthermore, BeLLM outperforms baselines in the average performance. 
These results indicate that incorporating backward dependencies to LLMs can helpfully learn sentence embeddings, which benefit downstream task performances.
%can improve sentence embeddings, ultimately leading to improved performance in downstream tasks.

\subsection{Ablation Study}
\label{exp::ablation}
We have shown the overall effectiveness of BeLLM. 
Here, we further analyze the effects of its varying settings with an ablation study results in Table \ref{table_ablation}.
% The first is the default turning point, set at $t=31$, selected based on the turning point guideline, as discussed in Section \ref{model::degradation}. At this point, BeLLM achieves the best performance. 
% The second turning point is $32$. It means the backward dependencies are not enhanced since LLaMA-7B only has $32$ attention layers. 
% Its result indicates that LLMs without enhanced backward dependencies perform worse than those with, demonstrating thehelpfultance of backward dependencies.
% The final turning point is $1$. It means all attention layers are bi-directional. 
From the \textit{turning point} ablation results, we find bi-directional attention layers are useful to BeLLM, yet converting all attention layers into bi-directional results in deficient performance.
It is possibly because uni-directional attention layers are crucial in language generation for representative word prediction.
%, as it utilizes prompt engineering to generate usual words.
Consequently, finding a good balance of uni- and bi-directional attention layers plays a crucial role in sentence embeddings.
%This ablation study highlights the importance of balancing the generation ability and backward dependencies for sentence embedding learning. 
% It also validates the guidelines of selecting the turning point.

% We then conduct another ablation study to compare two bi-directional strategies of introducing backward dependencies. 
% Table  \ref{table::ablation} 
From the \textit{bi-directional strategy} results, we observe that the ``modification'' strategy yields better results than the ``addition'' strategy. 
This implies that the increase in trainable parameters to LLMs by adding new attention layers will increase the difficulty of fine-tuning the model. 
It demonstrates that modifying the uni-directional attention layer by removing casual masks is effective.
%We thus adopt the ``modification'' strategy as the default strategy to introduce backward dependencies. 
%Nevertheless, the ``modification'' strategy is a practical approach to incorporating backward dependencies into LLMs. 

\begin{table}[ht]
\small
\centering
\begin{threeparttable}
\begin{tabular}{lc}
\toprule
Ablation Models &  Spearman's $\rho$
%Spearman's Correlation 
\\
\midrule
\midrule
\multicolumn{2}{c}{\textit{Turning Point}} \\
\midrule
BeLLM (\textit{last bi-layer})
%(t=31) 
(default) 
& $\mathbf{86.26}$ \\
BeLLM (\textit{no bi-layer})
%(t=32) 
& $85.24$ \\
BeLLM (\textit{all bi-layers})
%(t=1)  
& $75.55$ \\
\midrule
\midrule
\multicolumn{2}{c}{\textit{Bi-directional Strategy}} \\
\midrule
\textit{Modification} & $\mathbf{86.26}$ \\
\textit{Addition} & $84.35$ \\
\bottomrule
\end{tabular}
\end{threeparttable}
\caption{The results of BeLLM ablations with the average Spearman's correlation (Spearman's $\rho$) of S-STS.
\textit{Last bi-layer}: bi-directional layer at the last layer; no bi-layer and all bi-layers: the BeLLM ablation with no and all bi-directional layers.
\textit{Modification}: removing casual masks;
\textit{Addition}: adding new bi-directional layers.
}
\label{table_ablation}
\end{table}

\subsection{Case Study}\label{exp::case}
\begin{table}[ht]
\scriptsize
\begin{tabular}{ll}
\toprule
\multicolumn{1}{l}{\textbf{Q:}} & A \textcolor{applegreen}{female police officer} \textcolor{applegreen}{wears} an officer's \textcolor{applegreen}{hat} and \textcolor{applegreen}{sunglasses}.
\\ \midrule 
\multicolumn{2}{c}{SimCSE$_{RoBERTa}$}
\\ \midrule
\multicolumn{1}{l}{\#1}            & An officer stands next to a car on a city street.                                                                                                 \\ \midrule
\multicolumn{1}{l}{\#2}            & A \textcolor{applegreen}{police woman} smiling and \textcolor{applegreen}{wearing} \textcolor{applegreen}{sunglasses} and a \textcolor{applegreen}{hat}.                                                                                          \\ \midrule
\multicolumn{1}{l}{\multirow{2}{*}{\#3}} & \multicolumn{1}{l}{Young, smiling, blond \textcolor{applegreen}{female police officer} from New York}                                            \\ %\cline{2-2} 
\multicolumn{1}{l}{}                     & standing outside on a sidewalk. 
\\ \midrule
\multicolumn{1}{l}{\multirow{2}{*}{\#4}} & \multicolumn{1}{l}{An officer in a black uniform and \textcolor{applegreen}{hat} stands to the left of a}                                            \\ %\cline{2-2} 
\multicolumn{1}{l}{}                     & large structure with other officers in the background.
\\ \midrule
\multicolumn{2}{c}{SimCSE$_{LLaMA}$}                                                                                                                                                             \\ \midrule
\multicolumn{1}{l}{\multirow{2}{*}{\#1}} & \multicolumn{1}{l}{A \textcolor{applegreen}{female police officer} in a \textcolor{applegreen}{cap} and navy uniform smiles}                                             \\ %\cline{2-2} 
\multicolumn{1}{l}{}                     & while \textcolor{applegreen}{wearing sunglasses} outside of a shop.   \\ \midrule
\multicolumn{1}{l}{\multirow{2}{*}{\#2}} & \multicolumn{1}{l}{Young, smiling, blond \textcolor{applegreen}{female police officer} from New York}                                            \\ %\cline{2-2} 
\multicolumn{1}{l}{}                     & standing outside on a sidewalk. 
\\ \midrule
\multicolumn{1}{l}{\#3}            & An officer stands next to a car on a city street.                                                                                                 \\ \midrule
\multicolumn{1}{l}{\multirow{2}{*}{\#4}} & \multicolumn{1}{l}{An officer in a black uniform and \textcolor{applegreen}{hat} stands to the left of a}                                            \\ %\cline{2-2} 
\multicolumn{1}{l}{}                     & large structure with other officers in the background.
\\ \midrule
\multicolumn{2}{c}{BeLLM}                                                                                                                                                             \\ \midrule
\multicolumn{1}{l}{\multirow{2}{*}{\#1}} & \multicolumn{1}{l}{A \textcolor{applegreen}{female police officer} in a \textcolor{applegreen}{cap} and navy uniform smiles}                                             \\ %\cline{2-2} 
\multicolumn{1}{l}{}                     & while \textcolor{applegreen}{wearing} \textcolor{applegreen}{sunglasses} outside of a shop.
\\ \midrule
\multicolumn{1}{l}{\#2}            & A \textcolor{applegreen}{police woman} smiling and \textcolor{applegreen}{wearing} \textcolor{applegreen}{sunglasses} and a \textcolor{applegreen}{hat}.                                                                                          \\ \midrule
\multicolumn{1}{l}{\multirow{2}{*}{\#3}} & \multicolumn{1}{l}{Young, smiling, blond \textcolor{applegreen}{female police officer} from New York}                                            \\ %\cline{2-2} 
\multicolumn{1}{l}{}                     & standing outside on a sidewalk. 
\\ \midrule
\multicolumn{1}{l}{\multirow{2}{*}{\#4}} & \multicolumn{1}{l}{An attractive young New York City \textcolor{applegreen}{police woman} pauses}                                            \\ %\cline{2-2} 
\multicolumn{1}{l}{}                     & on the sidewalk.

\\ \bottomrule
\end{tabular}
\caption{
The top 4 retrieved sentences by RoBERTa, LLaMA, and BeLLM from the flickr30k dataset. The words with green color represent the keywords.
}
\label{table_case_study}
\end{table}

To better understand why BeLLM performs well,  we conduct a text retrieval experiment on the test split of the flickr30k dataset \citep{young-etal-2014-image}.
It contains images, each with 5 captions.
%to evaluate the sentence embedding quality. 
%This dataset includes five similar captions for each photo. 
We used the first caption vector to retrieve the top 4 similar sentences using the faiss vector search framework \citep{johnson2019billion}. 
For strict accuracy (correct cases only count for the top 4 retrieved captions exactly matching the 4 references), SimCSE$_{RoBERTa}$, SimCSE$_{LLaMA}$, and BeLLM obtains $16.4\%$, $18.4\%$, and $20.2\%$, respectively. It shows that BeLLM's sentence embeddings superiorly reflect semantic similarities for retrieval.
%}

%\textcolor{blue}{
We then examine a case in Table \ref{table_case_study}.
%to examine different retrieved captions.
%the quality of sentence embeddings. 
In the query caption, ``police officer'' is the subject, and ``female'' is its subject modifier; their relations can be learned through forward dependency; ``hat'' and ``sunglasses'' are objects, and their relations to ``police'' should be indicated by backward dependency.
SimCSE$_{RoBERTa}$ has a smaller model scale, which limits its ability to exploit global context. 
As a result, its top results even messed up the subject.
% retrieval quality is lower compared to larger-scale models. 
% For instance, it fails to include the correct subject and modifier in retrieved caption 1, and it also misses the correct objects in sentence 3.
Despite having a larger model scale, 
%and the ability to learn global context effectively, 
SimCSE$_{LLaMA}$ exhibits inferior results due to the lack of backward dependency modeling.
For this reason, most of the retrieved results missed 1-2 objects.
%suffers from poor retrieval quality due to its autoregressive architecture, which fails to capture backward dependencies. 
% For example, its second-ranked caption contains incorrect objects, and sentence 3 has wrong subjects and objects.
On the contrary, BeLLM performs the best, and the top two captions exactly match the references.
The results indicate the superiority of BeLLM in exploiting context to learn high-quality sentence embeddings.
% BeLLM outperforms SimCSE$_{RoBERTa}$ and SimCSE$_{LLaMA}$. Its top 1 and 2 retrieved sentences perfectly match the query, including the correct subject and object. This superior performance is due to BeLLM's strong ability to learn global contexts and capture both forward and backward dependencies.
%}

\subsection{Discussion of Enhanced Dependency}
\label{exp::discussion_enhanced}
\begin{figure}[ht]
    \centering
    \includegraphics[width=0.98\textwidth, trim = {0 0 0 10}, clip]{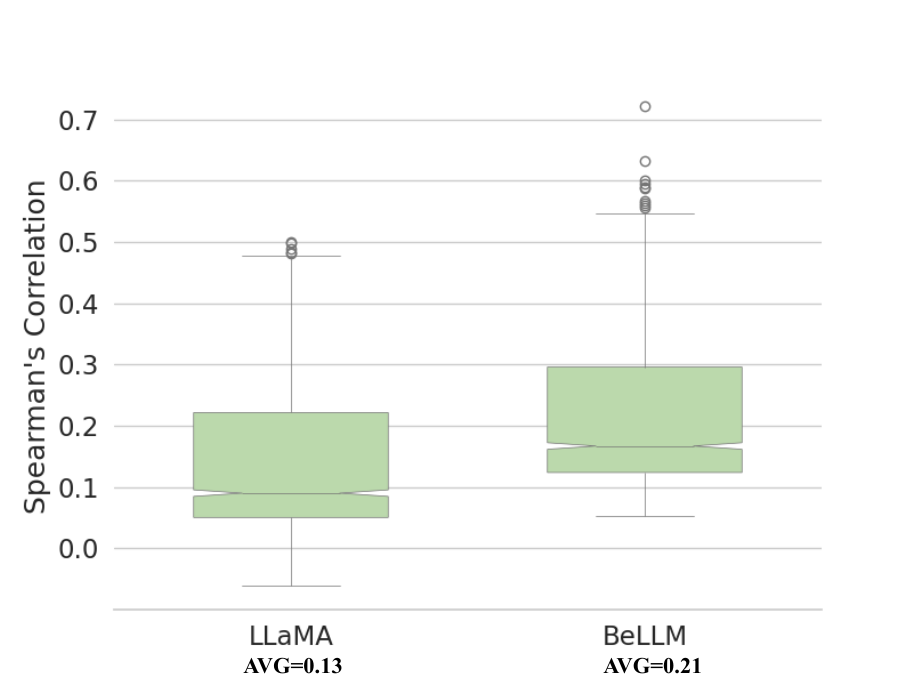}
    \caption{The sentence-level Spearman correlation box plot of LLaMA and BeLLM on the STS-B test set.}
    \label{figure_enhanced_dependencies}
\end{figure}

Finally, we experiment on the intra-sentence dependencies for BeLLM to supplement Section \ref{ssec:pilot-analysis}. 
Figure \ref{figure_enhanced_dependencies} depicts the dependencies changes from LLaMA to BeLLM. As can be seen, BeLLM exhibits much higher overall Spearman's correlation scores than LLaMA. This evidence again demonstrates that transforming the last attention layer bi-directional is effective in enhancing the capabilities of capturing dependency and is  helpful in exploiting sentence context for embeddings.

\section{Related Work}
%There are two main approaches for learning sentence embeddings: unsupervised and supervised.
BeLLM is in line with sentence embeddings. 
%\paragraph{Unsupervised Approaches} 
%Early studies \citep{DBLP:conf/naacl/HillCK16,pagliardini-etal-2018-unsupervised} have indicated the effectiveness of augmenting word2vec \citep{word2vec_mikolov_2013} with n-gram. 
In contrast to early attempts focusing on embeddings on the word level \citep{word2vec_mikolov_2013}, sentence embeddings capture sentence-level semantics to understand context better.
%However, they rely on static word embeddings and lack the ability to capture sentence-level semantic information when learning sentence embeddings.
Many previous studies adopted \textbf{unsupervised} approaches to utilize large-scale unlabeled text.
%To address this issue, 
Here, a popular method is to use BERT-alike transformers \citep{li-etal-2020-sentence,su2021whitening} to encode sentence embeddings. 
%However, these methods merely transform the distribution of BERT without fine-tuning it to achieve better performance.
%To tackle this drawback, 
Based on that, contrastive learning \citep{zhang-etal-2020-unsupervised, simcse_gao_2021, zhuo-etal-2023-whitenedcse, wu-etal-2023-hicl} was further employed to explore semantic similarities for sentence embeddings with self-supervision.
%have been proposed to improve sentence embeddings through contrastive fine-tuning.

To better align sentence embeddings to human senses, other prior studies \citep{conneau-etal-2017-supervised,cer-etal-2018-universal} employed labeled data for \textbf{supervised} sentence embedding learning.
Here, pre-trained language models typically worked as the backbone architectures \citep{sbert-nils-2019,cosent_su_2022}.
Following that, there is growing attention to the use of pre-trained LLMs for sentence embeddings \citep{li2023angle,jiang2023scaling}.  
%\paragraph{Supervised Approaches} 
%Many prior studies \citep{conneau-etal-2017-supervised,cer-etal-2018-universal} used supervised datasets to learn sentence embeddings. 
%Recently, \citet{sbert-nils-2019,cosent_su_2022} employed the pretrained language model to improve sentence embedding. 
%Very recently, \citet{li2023angle,jiang2023scaling} applied the LLMs to further improve sentence embeddings. 
%However, these LLMs-based sentence embeddings ignore the importance of backward dependencies.
However, existing works employed commonly used autoregressive LLMs, neglecting the potential benefits of backward dependencies to sentence embeddings.
Viewing this gap, we extensively explore the effects of backward dependencies in LLMs for sentence embeddings.

\section{Conclusion}
Our work has pointed out the benefits of coupling forward and backward dependencies in LLMs for sentence embeddings.
We have introduced BeLLM, a novel LLM with backward dependency-enhanced sentence embedding learning.
%backward dependency-enhanced large language model for improving sentence embeddings. 
%Our research unveils a turning point in the generation ability of LLMs for STS tasks. 
% Beyond this turning point, additional generation ability does not yield gains in STS performance.
% BeLLM converts the attention layers exceeding the turning point from uni- to bi-directional. 
% This transformation effectively enhances BeLLM's capacity to capture backward dependencies, resulting in improved performance in STS tasks.
In extensive experiments, BeLLM has achieved state-of-the-art performance across varying STS and downstream tasks.
%, highlighting its effectiveness in improving sentence embeddings.

\section*{Limitations}
BeLLM has large-scale parameters, which can hinder its efficiency when applied to real-world applications. Addressing this challenge is an essential aspect of our future work. We plan to optimize the model, ensuring it remains practical and efficient for real-world applications. 

\section*{Acknowledgements}
This work is supported by the NSFC Young Scientists Fund (Project No. 62006203), a grant from the Research Grants Council of the Hong Kong Special Administrative Region, China (Project No. PolyU/25200821), the Innovation and Technology Fund (Project No. PRP/047/22FX), and PolyU Internal Fund from RC-DSAI (Project No. 1-CE1E).

Here, we sincerely thank the reviewers and ACs for their valuable input, which has greatly improved our work.

\bibliography{anthology,custom}

\appendix
% \clearpage
\section{Discussion of Anisotropy Problem}
\label{sec:anisotropy}

Our proposed representative word strategy and contrastive learning can mitigate the anisotropy issue.

Firstly, our proposed representative word strategy can help alleviate this issue since representative words are typically not high-frequency words. For instance, in our experiment using the BeLLM model, the representative word for the sentence ``I am unhappy because it is raining'' is ``unhappy'', which is not as commonly used as words like ``it'' and ``is''. Our quantitative experiment in the STS-B dataset indicates that approximately $91.74\%$ of representative words are not high-frequency words.

Furthermore, the use of contrastive learning can alleviate this problem further. 
Here, we give a theoretical explanation. Following \citet{wang2020understanding}, the contrastive learning objective of Eq. \ref{eq_contrastive} can be formulated as follows:
\begin{equation}
    \label{eq_contrastive2}
    \small
    \begin{split}
        &\mathcal{L}_{contrastive} = \mathbb{E}_{(x, x^+) \sim p_{pos}} \left [ -f(x)^T f(x^+)/ \tau \right ] \\
        &+ \underset{{(x, x_i^-) \sim p_{data}}}{
            \underset{{(x, x^+) \sim p_{pos}}}{\mathbb{E}}
        } \left[ \mathrm{log} \left(
            e^{f(x)^Tf(x^+)/\tau} + \sum_i e^{f(x_i^-)^Tf(x)/\tau}
        \right) \right],
    \end{split}
\end{equation}
where $x^+$ and $x^-$ are the positive and negative examples of $x$, respectively. $p_{pos}$ denotes positive instances. 
$p_{data}$ is uniform over finite samples ${x_i}_{i=1}^m$. $f(x)^Tf(x^+)/\tau$ measures the alignment (similarity). 
Since the term $\sum e^{f(x_i^-)^Tf(x)/\tau}$ is always positive, the loss function inherently favors smaller values for $\mathbb{E} \left [ -f(x)^T f(x^+)/ \tau \right ]$, i.e., it can be optimized well. 
Suppose the encoder is perfectly aligned, i.e., $\mathbb{P}[f(x)=f(x^+)]=1$, then minimizing Eq. \ref{eq_contrastive2} equally minimizes following equation:
\begin{equation}
    \label{eq_contrastive3}
    \small
    \underset{{(x, x_i^-) \sim p_{data}}}{
            \underset{{(x, x^+) \sim p_{pos}}}{\mathbb{E}}
        } \left[ \mathrm{log} \left(
            e^{1/\tau} + \sum_i e^{f(x_i^-)^Tf(x)/\tau}
        \right) \right],
\end{equation}
To minimize it, all sentence embeddings should be pushed away from each other to form a \textit{uniform} distribution.
The uniform distribution of sentence embeddings can eliminate the effect of the common words and thus alleviate the anisotropy problem.

\section{Examples of CSTS}
\label{sec::csts_intro}

Table \ref{table_csts_example} shows an example of C-STS. Each sample in C-STS  includes three fields: sentence 1, sentence 2, and condition. The sentence pair exhibits varying similarities based on different conditions. The same sentence pair can be classified as either high similarity or low similarity under different conditions. The scale ranges from 1 (dissimilar) to 5 (similar).

% \usepackage[normalem]{ulem}
\begin{table}[ht]
\small
\begin{tabular}{ll}
\toprule
\multicolumn{2}{l}{\textbf{Sentence 1:} An older man holding a glass of wine} \\
\multicolumn{2}{l}{while standing between two beautiful ladies.}     \\ \midrule
\multicolumn{2}{l}{\textbf{Sentence 2:} A group of people gather around a}    \\ 
\multicolumn{2}{l}{table with bottles and glasses of wine.}         \\ \midrule
\multicolumn{1}{c|}{Condition}                    & Similarity       \\ \midrule
\multicolumn{1}{l|}{The people’s demeanor}        & 5                \\ \midrule
\multicolumn{1}{l|}{The number of bottles}        & 1                \\ \bottomrule
\end{tabular}
\caption{An example from the C-STS validation set, where the same sentence pair has different similarities based on different conditions. A similarity score of 1 indicates dissimilar, while a score of 5 represents similar.}
\label{table_csts_example}
\end{table}

\end{document}